\documentclass[conference]{IEEEtran}

\IEEEoverridecommandlockouts

\usepackage{cite}
\usepackage{amsmath,amssymb,amsfonts}
\usepackage{algorithmic}
\usepackage{graphicx}
\usepackage{textcomp}
\usepackage{xcolor}

\def\BibTeX{{\rm B\kern-.05em{\sc i\kern-.025em b}\kern-.08em
    T\kern-.1667em\lower.7ex\hbox{E}\kern-.125emX}}
\begin{document}

\title{Event Interval Modulation: A Novel Scheme \\
for Event-based Optical Camera Communication
}



\author{\IEEEauthorblockN{
    Miu Sumino\textsuperscript{*},
    Mayu Ishii\textsuperscript{*†},
    Shun Kaizu\textsuperscript{†},
    Daisuke Hisano\textsuperscript{‡},
    Yu Nakayama\textsuperscript{*}}

\IEEEauthorblockA{\textsuperscript{*}Department of Computer and Information Sciences, \\
Tokyo University of Agriculture and Technology, Tokyo, Japan\\
\{miu.sumino, mayu.ishii, yu.nakayama\}@ynlb.org}

\IEEEauthorblockA{\textsuperscript{†}Sony Semiconductor Solutions Corporation, Tokyo, Japan}

\IEEEauthorblockA{\textsuperscript{‡}Graduate School of Engineering, The University of Osaka, Osaka, Japan\\
hisano@comm.eng.osaka-u.ac.jp, hisano@ieee.org}
}

\maketitle

\begin{abstract}
Optical camera communication (OCC) represents a promising visible light communication technology. Nonetheless, typical OCC systems utilizing frame-based cameras are encumbered by limitations, including low bit rate and high processing load. To address these issues, OCC system utilizing an event-based vision sensor (EVS) as receivers have been proposed. The EVS enables high-speed, low-latency, and robust communication due to its asynchronous operation and high dynamic range. In existing event-based OCC systems, conventional modulation schemes such as on-off keying (OOK) and pulse position modulation have been applied, however, to the best of our knowledge, no modulation method has been proposed that fully exploits the unique characteristics of the EVS. This paper proposes a novel modulation scheme, called the event interval modulation (EIM) scheme, specifically designed for event-based OCC. EIM enables improvement in transmission speed by modulating information using the intervals between events. This paper proposes a theoretical model of EIM and conducts a proof-of-concept experiment. First, the parameters of the EVS are tuned and customized to optimize the frequency response specifically for EIM. Then, the maximum modulation order usable in EIM is determined experimentally. We conduct transmission experiments based on the obtained parameters. Finally, we report successful transmission at 28 kbps over 10 meters and 8.4 kbps over 50 meters in an indoor environment. This sets a new benchmark for bit rate in event-based OCC systems.
\end{abstract}


\section{Introduction}
Optical camera communication (OCC) is a promising technology for wireless communication in the future owing to the advantages including security, license, and cost-efficiency. The transmitter of OCC is a light source such as LED lights, LED panels, displays, or digital signage. The receiver is a complementary metal-oxide-semiconductor (CMOS) sensor in a camera~\cite{liu2024design}. One of the key advantages of OCC is that it does not interfere with other RF signals, such as WiFi and Bluetooth, as it generally operates within the visible light frequency band, utilizing wavelengths that do not overlap with those of other RF signals. This characteristic renders OCC suitable for deployment in aircraft and hospitals, where restrictions on RF wave usage are in place. It is also expected to be used in situations where it is desirable to have more means of communication (e.g. in disaster areas).
Moreover, the employment of OCC is also a possibility in RF-unreachable areas, including underwater environments~\cite{shigenawa2022predictive}.

A typical OCC utilizes a camera as a receiver, thereby giving rise to a number of challenges. Improving the transmission capacity remains a challenging task, as it is constrained by the frame rate of the camera. The sheer volume of data captured by cameras necessitates substantial computational resources for real-time processing. Additionally, cameras are susceptible to ambient light (sun and lighting)~\cite{liu2020some}, which can compromise communication stability.
The resolution of these issues has been proposed through the utilization of OCC employing an event-based vision sensor (EVS) as the receiver~\cite{wang2022smart}.
Note that in the following sections, OCC systems that employ EVS as the receiver are referred to as event-based OCC.
The EVS is distinguished from conventional frame-based cameras by its asynchronous recording of scene intensity changes~\cite{finateu2020}.
This provides the following benefits:
\begin{itemize}
    \item EVS records intensity changes asynchronously, enabling high-speed communication independent of the frame rate. In other words, it effectively captures the derivative of the transmitted signal.
    \item As EVS records only the necessary data, the amount of data is small and communication is possible with a low processing load.
    \item The high dynamic range of EVS makes them less sensitive to ambient light.
\end{itemize}
However, there are limitations to the speed and stability of the communication as little consideration has been given to coding schemes.
This paper proposes a new modulation method, event interval modulation (EIM) and its demodulation method, which exploits the characteristics of EVS.
The main contributions of this work are:
\begin{enumerate}
    \item To achieve efficient data transmission suited to the asynchronous nature of EVS, we propose EIM, a modulation scheme that maximally exploits the EVS’s ability to capture temporal intensity changes. This paper describes the principles and mechanisms of both modulation and demodulation in detail.
    \item We conduct OCC transmission experiments in an indoor environment to evaluate the communication performance. The proposed method is assessed in terms of transmission capacity and bit error rate.
\end{enumerate}

\section{Related Work}\label{sec:rltd}
\subsection{Optical Camera Communication}\label{sec:occ}

Higher order modulation techniques are an important issue in OCC to increase transmission capacity. The main modulation techniques for OCC include on-off keying (OOK), frequency shift keying (FSK), pulse amplitude modulation (PAM) and pulse width modulation (PWM)~\cite{saeed2019optical}. The low frame rates of commercial cameras, typically 30-60 FPS, cause flickering~\cite{chia2023high}. To overcome this problem, modulation schemes using undersampling~\cite{luo2018undersampled} and rolling shutter-based modulation~\cite{rachim2018multilevel} have been proposed. These require a significant amount of signal processing on the receiver side, as the target pixel must be continuously sampled on a frame rate basis.

\subsection{Event-based OCC}\label{sec:evs_occ}

EVS is characterized by each pixel operating independently and detecting changes in brightness, resulting in high temporal resolution and low latency~\cite{gallego2020event}. In ~\cite{lichtsteiner2008128}, the initial design of the event cameras and their high dynamic range and low latency characteristics are described in detail. Research on OCC using EVS has attracted much attention in recent years. In ~\cite{wang2022smart}, OCC was achieved using high-frequency modulated signals from LEDs and the event camera was used to demodulate the signals, achieving error-free communication of up to 4 kbps indoors and 500 bps outdoors over 100 meters. In ~\cite{nakagawa2024linking}, a study was conducted to link visual identification and communication between multiple agents via visible light communication using an event camera, which showed superior performance compared to conventional CMOS cameras.
These studies highlight the advantages of fast communication and low latency due to the asynchronous recording of EVS instead of conventional frame-based cameras. The previous works applied only conventional modulation schemes, but further improvements in transmission rate and robustness can be expected by leveraging the unique characteristics of EVS. This paper proposes EIM and conducts experimental validation.

\section{Event-based Vision Sensor}\label{sec:evs}

\begin{figure}[!t]
\centering
	\includegraphics[width=2.6in]{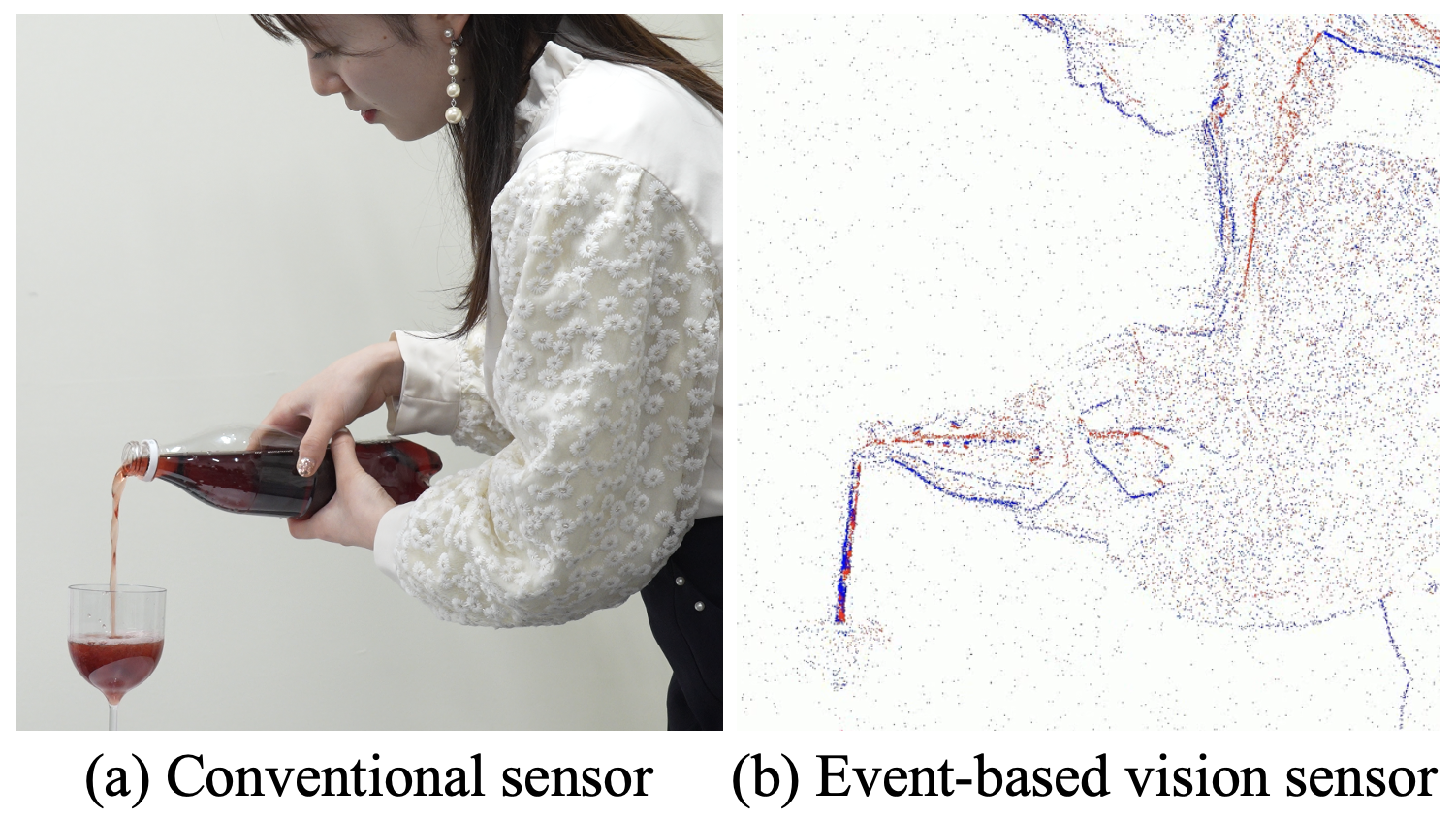}
	\caption{EVS photography.}
	\label{fig:image}
\end{figure}

\begin{figure}[!t]
\centering
	\includegraphics[width=3.3in]{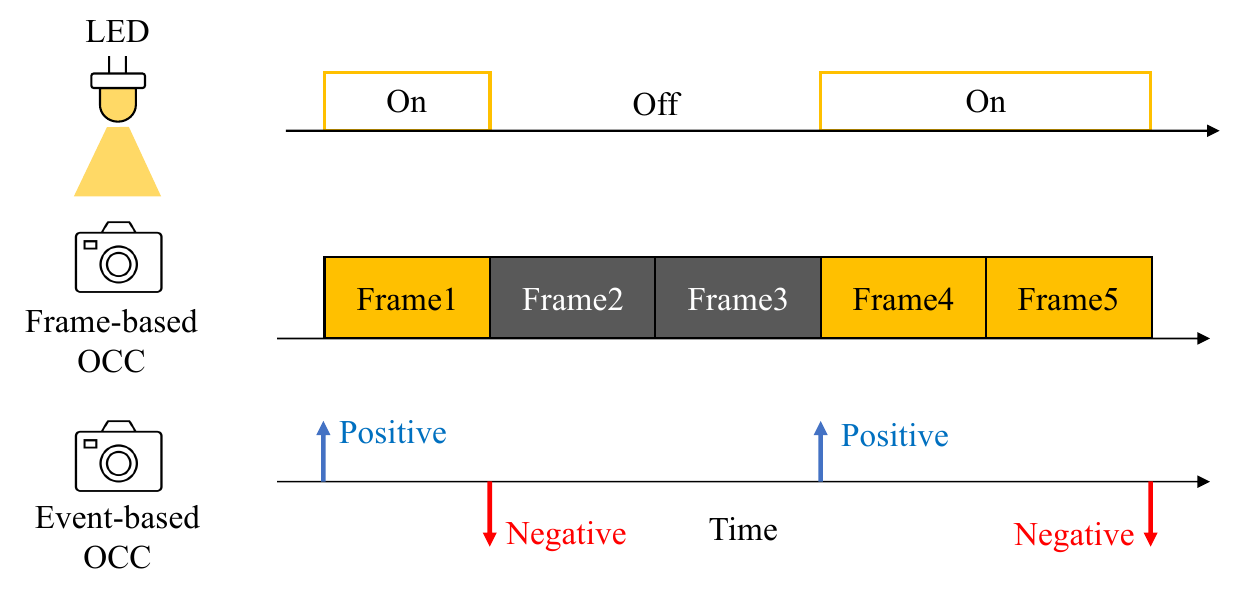}
	\caption{Received signal shapes of frame-based OCC and event-based OCC.}
	\label{fig:frame_event}
\end{figure}

In contrast to conventional frame-based cameras, the EVS is a sensor that asynchronously records changes in scene luminance and is characterized by its ability to capture changes in light at high speed.
Figure~\ref{fig:image}(a) shows a scene captured with a conventional RGB camera and Fig.~\ref{fig:image}(b) shows the same scene captured with an EVS.
The difference in the received signal shapes between frame-based OCC and event-based OCC using EVS is illustrated in Fig.~\ref{fig:frame_event}. In the frame-based approach, the on/off state of the signal is determined for each frame. In comparison, the event-based method identifies state transitions using positive and negative events triggered at rising and falling signal edges, respectively.
This section describes the operating principles of EVS.

\subsection{Pixel Structure and Parameters
}\label{sec:pixel_arch}

OCC demands the precise acquisition of on/off signals (1 kHz or higher) from rapidly blinking LEDs. In this application, the frequency of the events to be acquired is significantly higher than in the standard EVS use case. Consequently, specific adjustments are implemented to ensure precise capture of rapid signal transitions during optical camera communications.

\begin{figure}[!t]
\centering
	\includegraphics[width=2.5in]{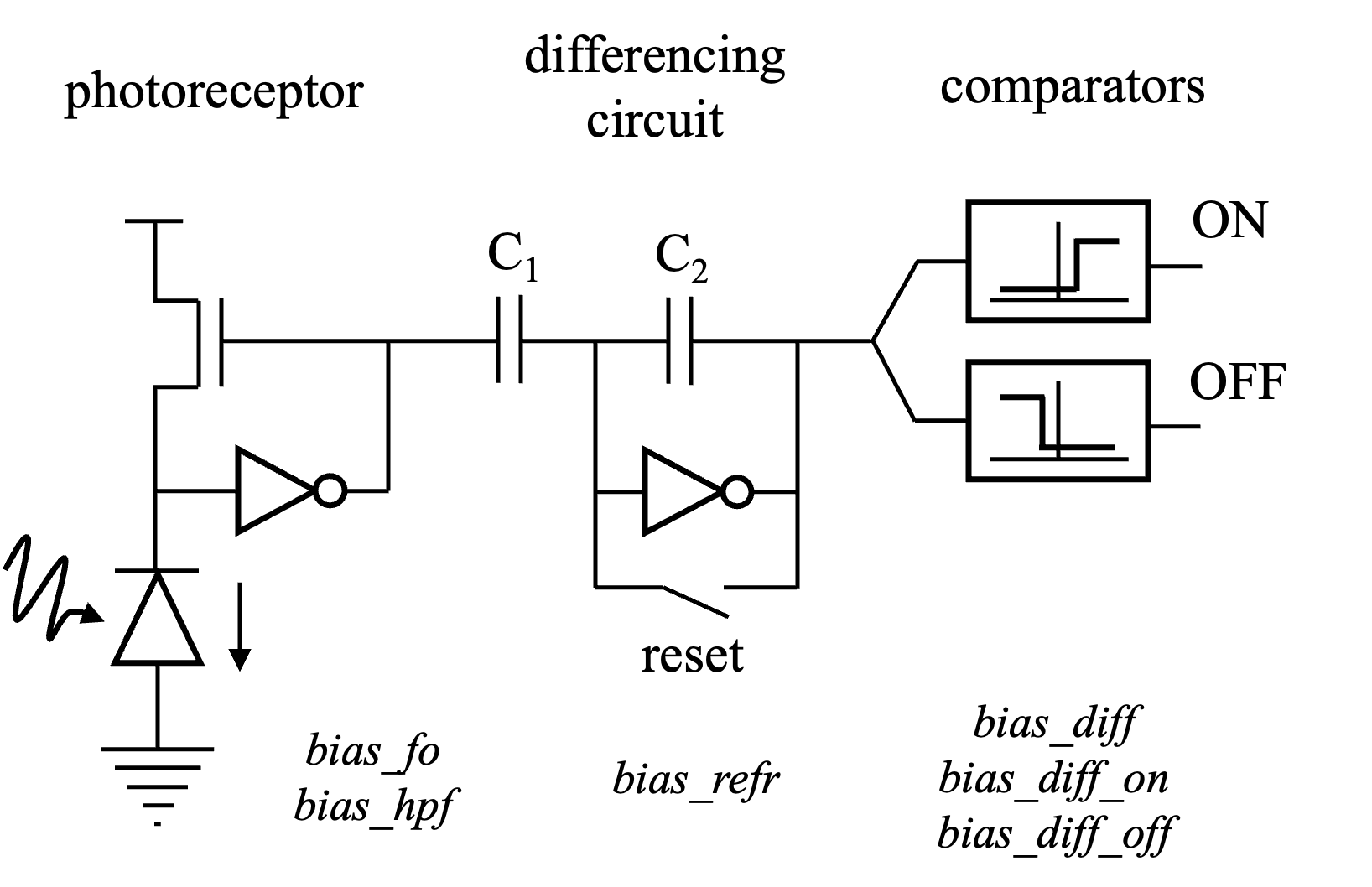}
	\caption{Abstract EVS pixel architecture.}
	\label{fig:arch}
\end{figure}

Figure~\ref{fig:arch} shows the pixel architecture of the EVS demonstrated in~\cite{lichtsteiner2008128}.
The light received by the photoreceptor is I-V converted and compared by the differencing circuit, which takes the difference between the previously stored brightness and the current brightness. If the difference is greater or less than a threshold, an event is output and the current brightness is stored.

We present the tunable parameters for fast response in the EVS used in this paper. For detail, see~\cite{PropheseeMetavisionSDK}. The low-pass filter ($\mathit{bias\_fo}$) and high-pass filter ($\mathit{bias\_hpf}$) can be configured as pixel bandwidth bias settings. The dead time bias ($\mathit{bias\_refr}$) controls the refractory period of the pixel. The contrast sensitivity threshold biases ($\mathit{bias\_diff}$, $\mathit{bias\_diff\_on}$, $\mathit{bias\_diff\_off}$) control the sensor’s sensitivity to changes in contrast.

\subsection{Response Characteristics of EVS}\label{sec:resp_evs}

The response characteristics of the pixel have been demonstrated to exert a significant influence on the events that can be obtained for LED on/off. In this study, an efficient and stable communication code is investigated by taking this characteristic into account.
Here we describe the analog response characteristics of EVS pixels for our coding considerations. In the context of OCC, the transient response of the pixel must be taken into account because the LEDs are switched on and off at high speed.

\begin{figure}[!t]
\centering
	\includegraphics[width=2.7in]{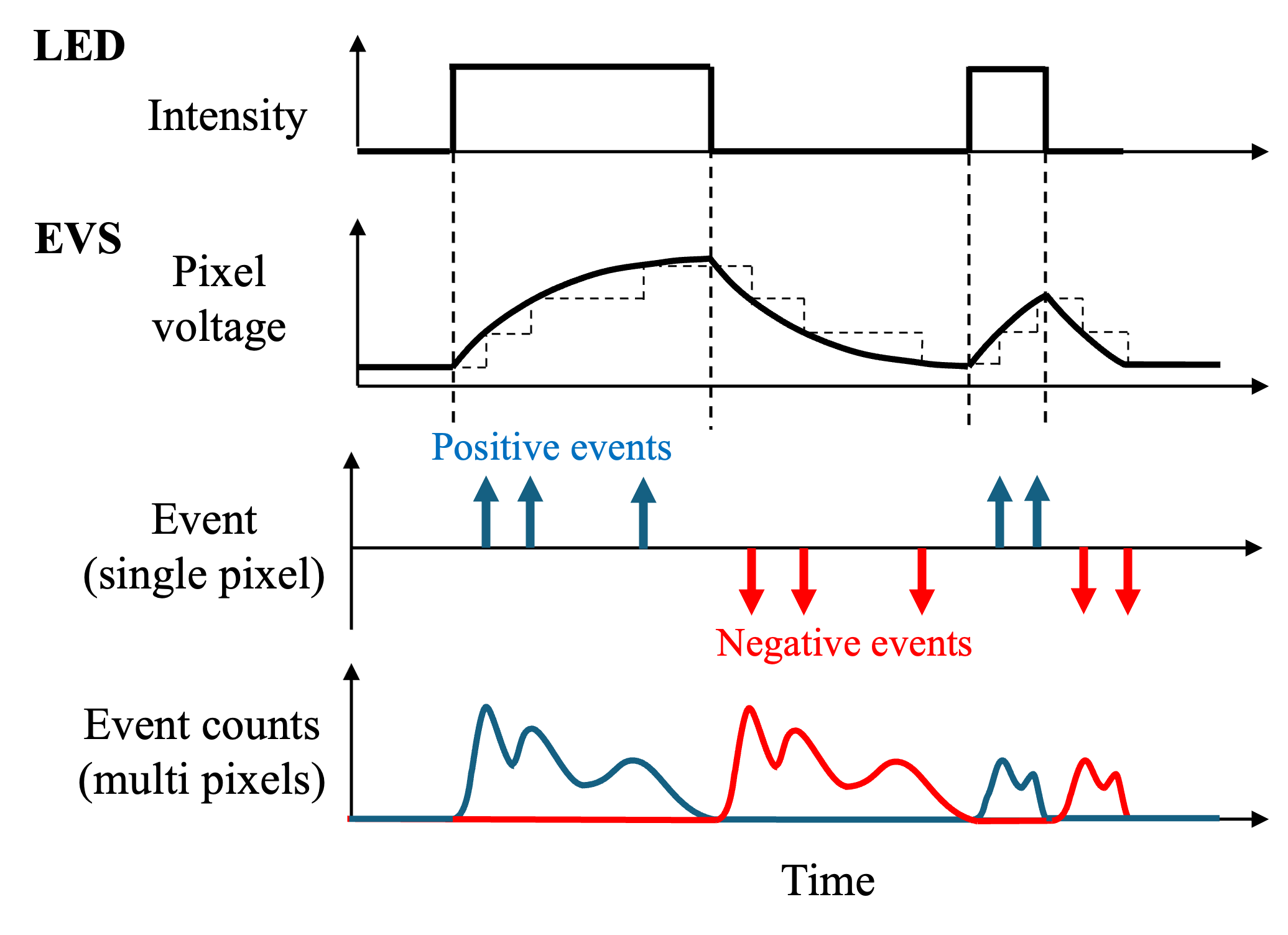}
	\caption{Event response to LED emission.}
	\label{fig:resp}
\end{figure}

As demonstrated in Fig.~\ref{fig:resp}, the figure of events generated in response to LED luminance changes is illustrated. In the instance of the LED transitioning from an off to an on state, the process of photoelectricity is converted by the photodiode, with the capacitor undergoing a state of charge. The occurrence of an event is initiated each time the threshold is surpassed during this transient response. Given the variability in event generation from pixel to pixel, a relatively continuous start-up waveform can be obtained by calculating the sum of the number of events for the pixels detecting the LED.

\section{Proposed EIM Scheme}\label{sec:pro}

Given that the pixel acquisition method of EVS has different characteristics from conventional CMOS image sensors, designing a new modulation scheme optimized for the requirements of EVS is expected to further improve transmission rate and robustness. This paper therefore evaluates the characteristics of EVS through experiments and proposes EIM, a modulation method suitable for these characteristics.

\subsection{Modulation}\label{sec:modulation}

A highly efficient modulation scheme that exploits the characteristics of EVS. The following describes the operating principle of EIM.

\subsubsection{General Formulation}\label{sec:general_mod}

This section describes the fundamental principles of EIM.
The number of event time intervals utilized for modulation is defined as $M\in {\mathbb N}_{\geq 2}$.
In the case where $M$ is a power of $2$, the quantity of information that can be transmitted for each event time interval is equal to $\log_2 M$ bits.
Conversely, when $M$ is not a power of $2$, it becomes necessary to combine multiple event time intervals and allocate bits so that the fractional part of $\log_2 M$ [bit] is as small as possible.
The following is a generalized description of these processes.
The number of event time intervals to be combined is denoted by $N$.
In this case, the optimal value of $N$, denoted by \(N^*\), is illustrated below:
\begin{equation}
  N^* = \operatorname*{arg\,min}_{N}
  \Bigl\{
    N \log_2 M
    \;-\;
    \lfloor N \log_2 M \rfloor
  \Bigr\}.
  \label{eq:optimal_n}
\end{equation}
\(\lfloor \cdot \rfloor\) is defined as the floor function.
The quantity of bits that can be allocated \(L\) is shown as follows:
\begin{equation}
  L = \Bigl\lfloor N^* \,\log_{2} M \Bigr\rfloor
  \label{eq:l}
\end{equation}
The set of available event time intervals is $T \;\equiv\;\{\,\tau_i \mid i \in \{1,\dots,M\}\}$.
The combination of event time intervals at $N^*$ is as follows:
\begin{equation}
  T^{(N^*)}
  \;=\;
  \bigl\{
    (\tau_{i_1},\,\tau_{i_2},\,\dots,\,\tau_{i_{N^*}})
    \;\bigm|\;
    i_k \in \{1,\dots,M\}
  \bigr\}
\end{equation}
Here, $\bigl|T^{(N^*)}\bigr| = M^{N^*}$.
From this direct product $T^{(N^*)}$, it is possible to select $2^L$ combinations of event time intervals and allocate $L$ bits.
The selection of event time interval combinations is based on the time widths that can be transmitted in the shortest possible time.
The set of time widths subsequent to optimization is denoted by $\Delta \;\subset\; T^{(N^*)}, \quad \bigl|\Delta\bigr| \;=\; 2^L$.
In other words:
\begin{equation}
  \Delta
  \;\equiv\;
  \bigl\{
    \delta_k
    \,\bigm|\,
    k \in \{1,\dots,2^L\},\;\delta_k \in T^{(N^*)}
  \bigr\}
\end{equation}
Here, we define $\delta_k
  \;=\;
  \bigl(\tau_{i_1}^{(k)},\,\tau_{i_2}^{(k)},\,\dots,\,\tau_{i_{N^*}}^{(k)}\bigr)
$.
\(\tau_{i_1}^{(k)}\) denotes the time width at the $k$-th selected element.
The total time \(d_k\) that is required to transmit this combination is defined as follows:
\begin{equation}
  d_k
  \equiv
  \sum_{j=1}^{N^*} \tau_{i_j}^{(k)}.
\end{equation}
The bit rate \(B\) [bps] is expressed as follows:
\begin{equation}
  B
  \;=\;
  \frac{2^L \,L}{\displaystyle \sum_{k=1}^{2^L} d_k}
\end{equation}

\subsubsection{Example: $M=2^L$, $L\in\mathbb{N}$}\label{sec:ex1}

An example is provided for the simple case of the general formula in Section~\ref{sec:general_mod}.
If $M$ is a power of $2$, the computation is greatly simplified.
When $M=2^L$ and $L\in\mathbb{N}$, $N^*=1$.
Furthermore, it is unnecessary to combine event time intervals, resulting in the equation $|T| = |\Delta|$.
In this case, the bit rate $B$ [bps] is as follows:
\begin{equation}
    B = \frac{ML}{\sum_{i=1}^{M} \tau_i}
    \label{eq:eq_ex1}
\end{equation}

\begin{figure}[!t]
\centering
	\includegraphics[width=3.0in]{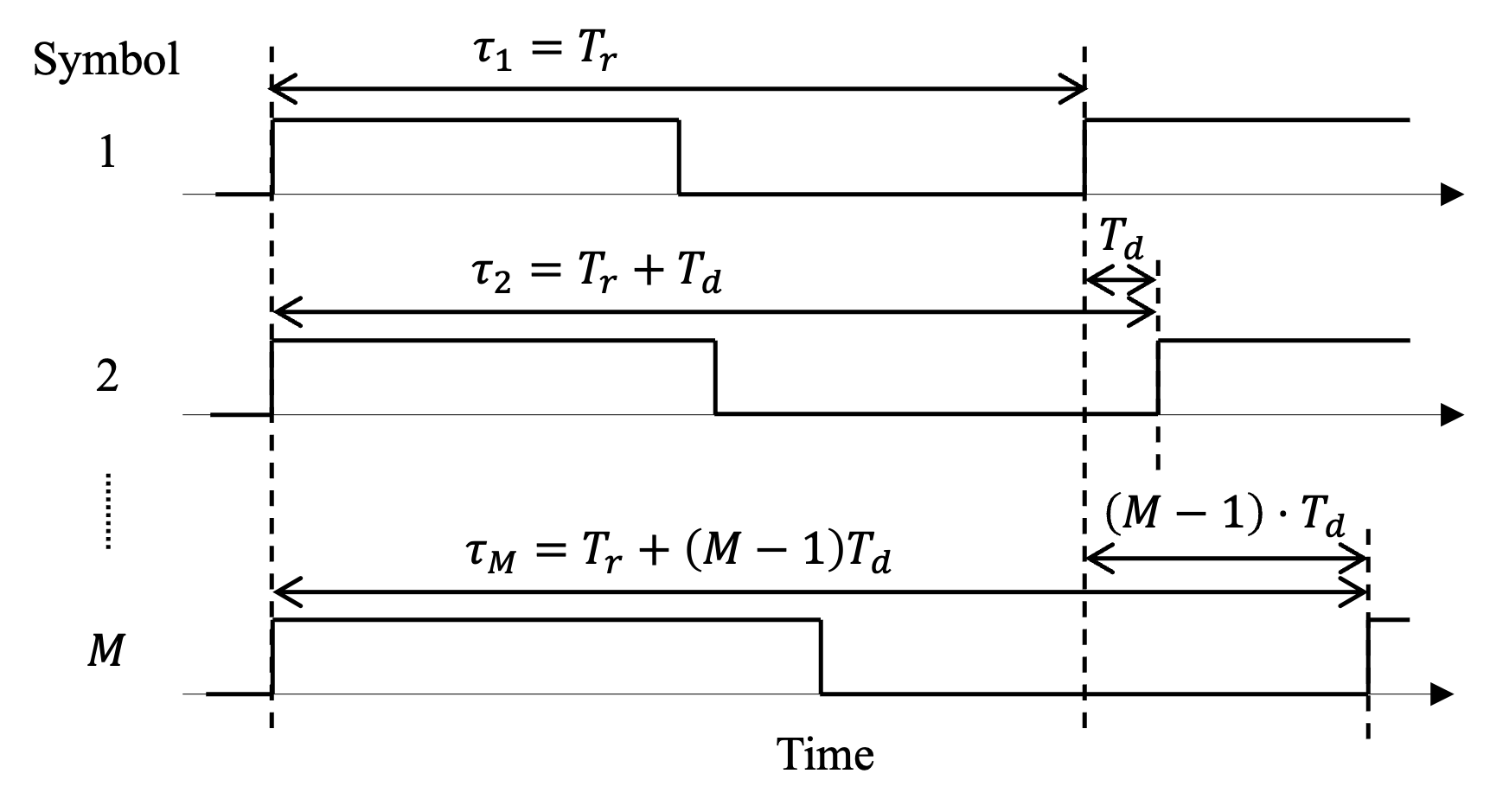}
	\caption{Event time interval defined by (\ref{eq:eq_ex2}).}
	\label{fig:eqex2}
\end{figure}

For the purposes of simplicity, the event temporal interval is defined in the following manner:
\begin{equation}
    \tau_i = T_{r}+(i-1)T_{d}
    \label{eq:eq_ex2}
\end{equation}
Each $\tau_i$ in M symbols is represented as in Fig.~\ref{fig:eqex2}. $T_{d}$ is defined as the minimum emission duration of the event. $T_{r}$ is the relaxation time required to quench the event, independent of the emission duration.
It should be noted that, in practice, the event emission durations do not align at the same discrete intervals as those presented in (\ref{eq:eq_ex2}).
The bit rate is as follows derived from (\ref{eq:eq_ex1}):
\begin{equation}
    B = \frac{2L}{2T_{r}+T_{d}(M-1)}
    \label{eq:bitrate}
\end{equation}

\subsection{Demodulation}\label{sec:demodulation}

This section describes the demodulation process in EIM.
First, upon the occurrence of an event in the EVS, the timestamp is recorded along with the number of positive and negative events. These event counts are passed through a smoothing filter to generate the received signals $r_p(t)$ and $r_n(t)$, corresponding to the positive and negative polarities, respectively. When no events occur, both signals are set to zero, i.e.,
$r_p(t) = 0$, $r_n(t) = 0$.
The demodulation process proceeds as follows:
\begin{enumerate}
    \item Detection of Negative Events: When the negative-side received signal $r_n(t)$ exceeds a predefined threshold $r_n^{\mathrm{(th)}}$, the negative event flag is set: $r_n(t) > r_n^{\mathrm{(th)}}$
    \item Detection of Transition to Positive Events: When events transition from negative to positive after the flag is set, the negative event flag is cleared.
    \item Detection of the First Local Maximum in Positive Periods: In periods where the negative event flag is not set (i.e., during positive event intervals), the first local maximum of the positive-side received signal $r_p(t)$ is detected. The time of this maximum is denoted as $t_p[l]$ and is defined as:
    \begin{equation}
      \begin{split}
      t_p[l] = \min \Bigg\{ t_{l-1} < t < t_l \ \Bigg| \
      \frac{d}{dt} r_p(t) = 0, \\
      \frac{d^2}{dt^2} r_p(t) < 0 \Bigg\}
      \end{split}
      \label{eq:local_max}
    \end{equation}
    where $t_{l-1}$ and $t_l$ represent the timestamps when the negative flag was set at $(l-1)$-th and $(l)$-th time, respectively.
    \item Measurement of Time Intervals: The time $t_p[l+1]$ corresponding to the next symbol is similarly obtained, and the symbol interval $t_d[l]$ is computed as:
    \begin{equation}
        t_d[l] = t_p[l+1] - t_p[l].
        \label{eq:interval}
    \end{equation}
    \item Selection of the Closest Event Interval: From a predefined set of event time intervals $T$, the element closest to $t_d[l]$ is selected as:
\begin{equation}
    \tau[l] = \operatorname*{arg\,min}_{\tau_i \in T} \left| t_d[l] - \tau_i \right|.
    \label{eq:select}
\end{equation}
    \item Symbol Decision Based on the Set $\Delta$: Finally, a symbol is determined by selecting the element in a predefined set $\Delta$ that is closest to the combination of the chosen interval $\tau[l]$.
\end{enumerate}

\section{Experiment}\label{sec:exp}
\subsection{Experimental Setup}\label{sec:setup}

\begin{figure}[!t]
\centering
	\includegraphics[width=2.6in]{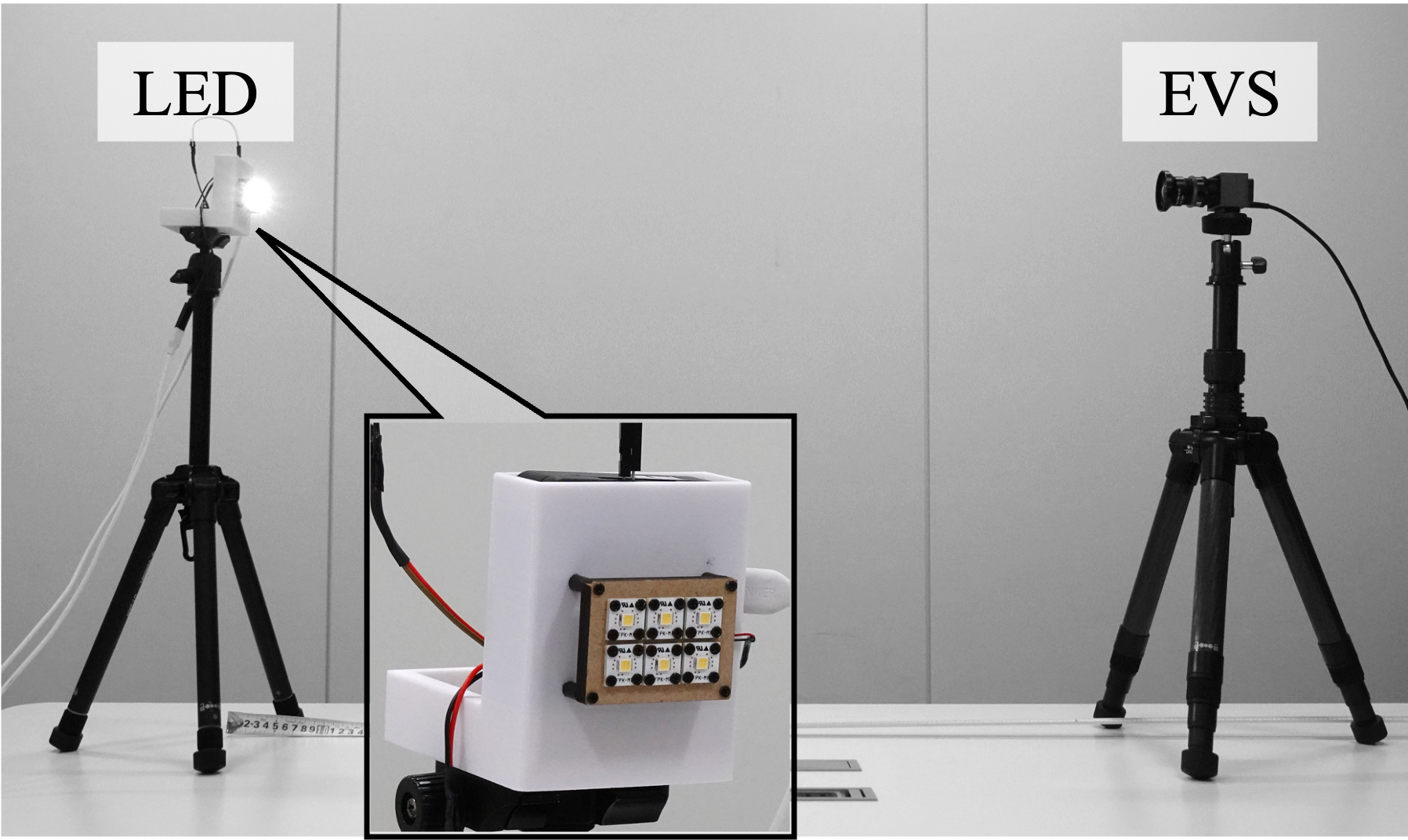}
	\caption{Experimental setup.}
	\label{fig:exp}
\end{figure}

\begin{figure}[!t]
\centering
	\includegraphics[width=2.8in]{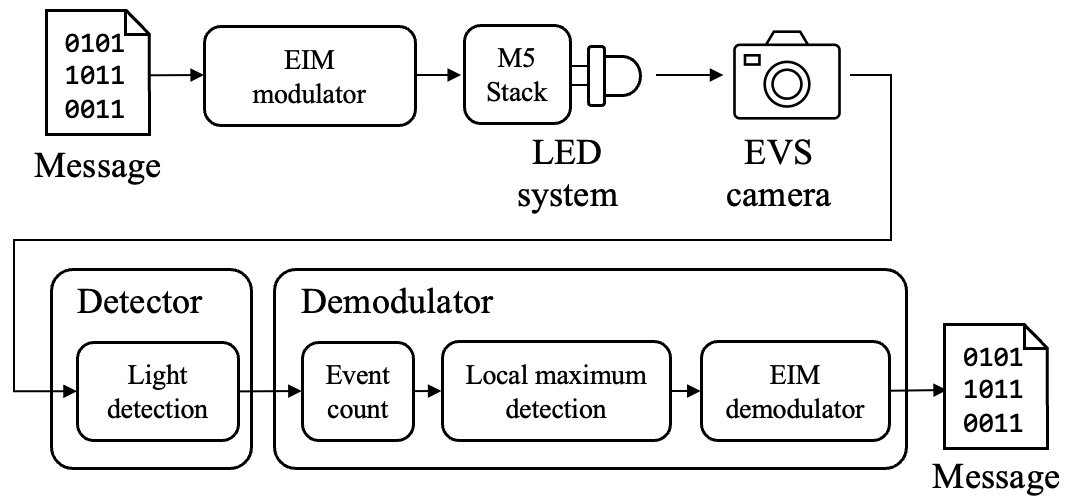}
	\caption{Experimental flow.}
	\label{fig:concept}
\end{figure}

Figure~\ref{fig:exp} shows the experimental setup. Figure~\ref{fig:concept} shows the functional blocks in this experiment. We generated a bitstream consisting of $10^6$ random symbols. The M5Stack modulated these symbols based on EIM and controlled the blinking of the LED. This experiment utilized surface mounting chip LEDs manufactured as the transmitter. The LED is white in color and has a brightness of 26.5 lm. In the demodulator, the EVS counted the number of events. Afterward, the offline digital signal processing (DSP) applied a smoothing filter, and calculated the intervals between events. Subsequently, EIM demodulation was performed, and the original bit sequence was recovered.

We conducted the experiment indoors, with an ambient light illuminance of 250 lux. This paper enhanced the characteristics of the EVS to support OCC systems. First, we carried out adjustment for the frequency response. Then, the optimal event interval was measured for each transmission distance. Based on these parameters, transmission experiments were subsequently performed.

\subsection{Adjustment for Frequency Response}\label{sec:para_tu}

\begin{figure}[!t]
\centering
	\includegraphics[width=3.4in]{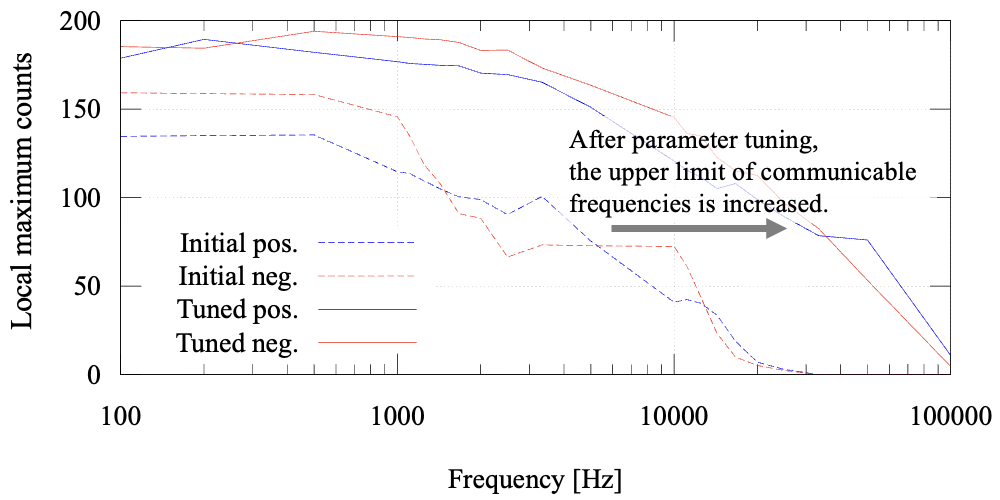}
	\caption{Change in number of events detected with initial and adjusted EVS parameters.}
	\label{fig:tuning}
\end{figure}

We calibrated each parameter of the EVS. The frequency characteristics of the EVS were measured by varying the on/off frequency of the LED. Figure~\ref{fig:tuning} shows the number of events observed at each frequency. Parameter tuning led to improved event occurrence at high frequencies above 20 kHz for both positive and negative signals, compared to the initial settings.

\subsection{Symbol Design}\label{sec:symbol}

The EIM parameters are defined as $T_r$, $T_d$ and the number of symbols $M$ in (\ref{eq:eq_ex2}).
This paper introduces two types of EIM parameters.
The first of these, Modulation 1, was optimized at a distance of 1 m for short-range, and Modulation 2 was optimized at a distance of 10 m for long-range.
This section explains the EIM parameter settings and the results.

\subsubsection{Setting of $T_r$}\label{sec:tr}

This section details the methodology of adjusting $T_r$, which denotes the minimum permissible pulse width.
During the modification of the on/off period of the LED, the EVS acquires the events that occur and the event cycle is calculated using the process outlined in Section~\ref{sec:demodulation}.
Based on the detection results of more than $10^5$ times, $T_r$ was set as the shortest on/off period that can be detected stably.
In consequence of the empirical measurements, $T_r$ in Modulation 1 was set to 32 $\mu$s and $T_r$ in Modulation 2 was set to 160 $\mu$s.

\subsubsection{Setting of $T_d$}\label{sec:td}

\begin{figure}[!t]
\centering
	\includegraphics[width=3.4in]{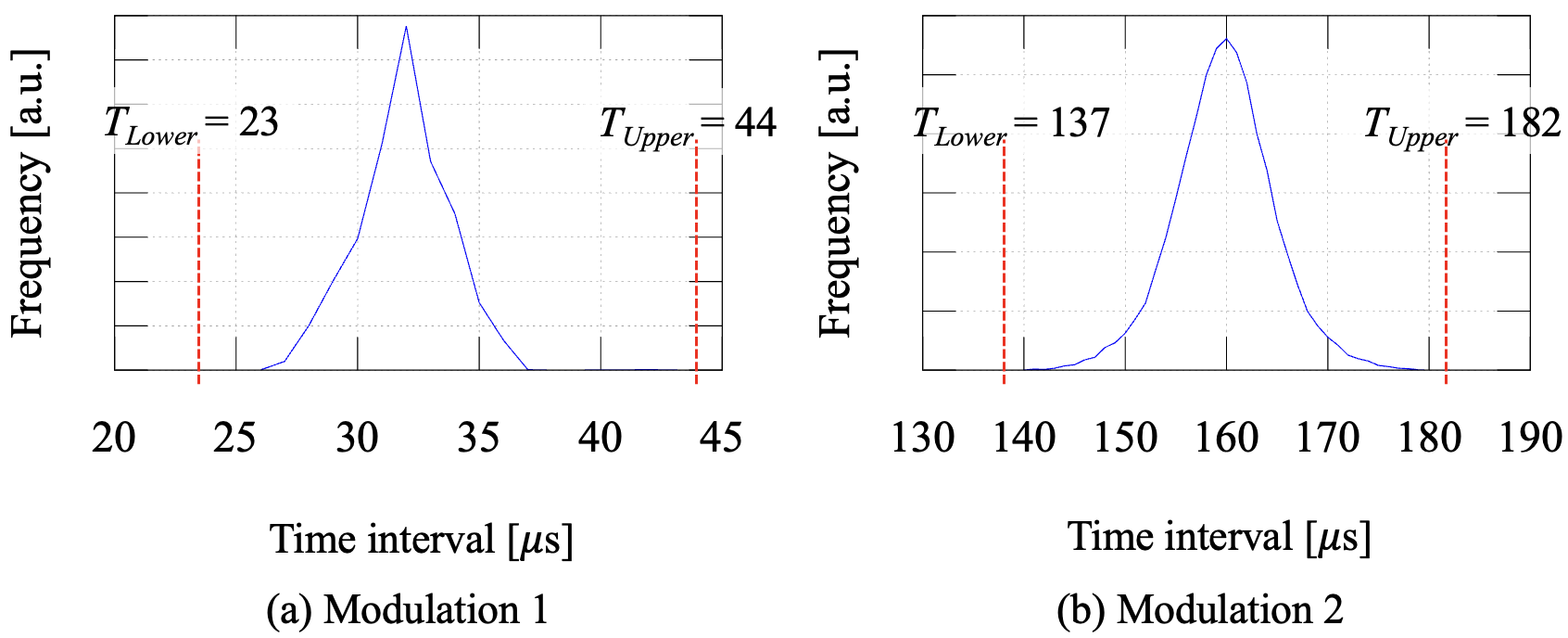}
	\caption{Histogram of the received time interval frequency.}
	\label{fig:hist}
\end{figure}

$T_d$ is the step amount of the pulse width. To prevent symbol errors during communication, the step amount $T_d$ must be set larger than the variation observed within each symbol. A histogram of the received time intervals was produced. Figure~\ref{fig:hist}(a) was carried out for Modulation 1 and Fig.~\ref{fig:hist}(b) for Modulation 2.
If the maximum detected time interval is designated $T_{Upper}$ and the minimum is $T_{Lower}$, it is necessary to ensure that $T_d > T_{Upper} - T_{Lower}$.
The $T_d$ in Modulation 1 was set to 26 $\mu$s and the $T_d$ in Modulation 2 was set to 60 $\mu$s as a result of the actual measurements.

\subsubsection{Setting of $M$}\label{sec:m}

\begin{figure}[!t]
\centering
	\includegraphics[width=3.4in]{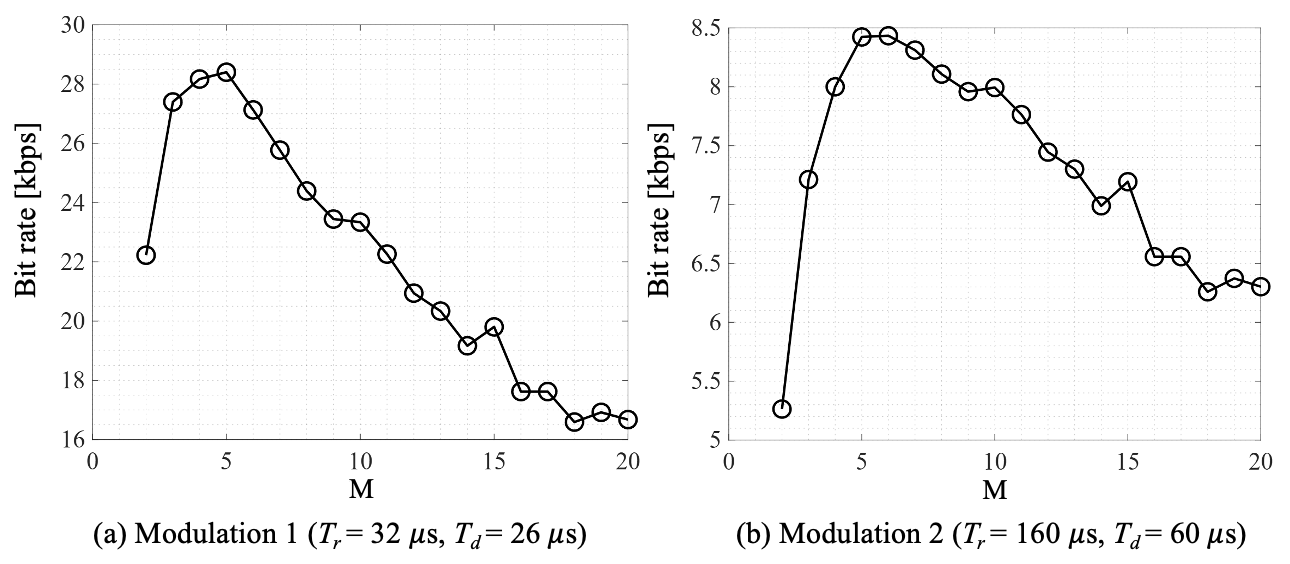}
	\caption{Number of symbols ($M$) vs. bit rate.}
	\label{fig:m_vs_bitrate_exp}
\end{figure}

The $T_r$ and $T_d$ values obtained above are utilized to ascertain the number of symbols $M$ for which the bit rate ($B$ in (\ref{eq:bitrate})) is maximum.
In this study, the search range for \(N^*\) by (\ref{eq:optimal_n}) was set to $N<10$.
The bit rate at each $M$ was calculated using (\ref{eq:optimal_n}), (\ref{eq:l}) and (\ref{eq:bitrate}). The number of symbols $M$ versus bit rate is shown in Fig.~\ref{fig:m_vs_bitrate_exp}.

As illustrated in Fig.~\ref{fig:m_vs_bitrate_exp}(a), when $M$ is set to maximize the bit rate, $M$ = 5 is optimal for short-range. The results in Fig.~\ref{fig:m_vs_bitrate_exp}(b) demonstrate that $M$=5 or $M$=6 is optimal for long-range.
In this study, considering the simplicity of modulation, a smaller $N^*$ was selected if the equivalent speed could be achieved. $M=4$ was selected for short-range, and $M=6$ was selected for long-range.
In summary, the EIM parameters that have been optimized are shown in Table~\ref{tbl:param}.

\begin{table}[!t]
  \caption{Modulation parameters.}
  \label{tbl:param}
  \centering
  \begin{tabular}{|c|c|c|}
    \hline
    \textbf{Item} & \textbf{Modulation1} & \textbf{Modulation2} \\
    \hline
    $T_r$ & 32 $\mu$s & 160 $\mu$s \\
    $T_d$ & 26 $\mu$s & 60 $\mu$s \\
    $M$  & 4               & 6               \\
    \(N^*\)  & 1               & 7 \\
    Bit rate & 28.2 kbps & 8.4 kbps \\
  \hline
  \end{tabular}
\end{table}

\subsection{Results}\label{sec:results}
\begin{figure}[!t]
\centering
	\includegraphics[width=3.4in]{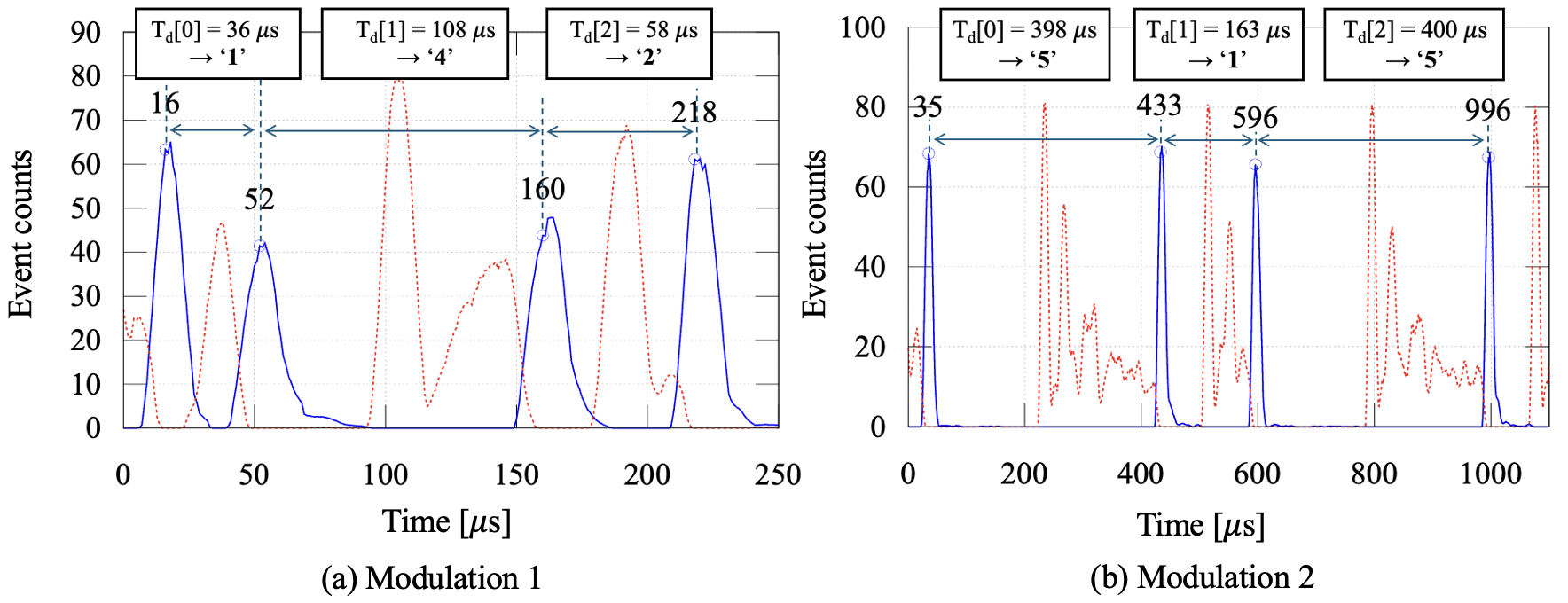}
	\caption{The EVS output and demodulation results.}
	\label{fig:peak_exp}
\end{figure}

\begin{figure}[!t]
\centering
	\includegraphics[width=3.0in]{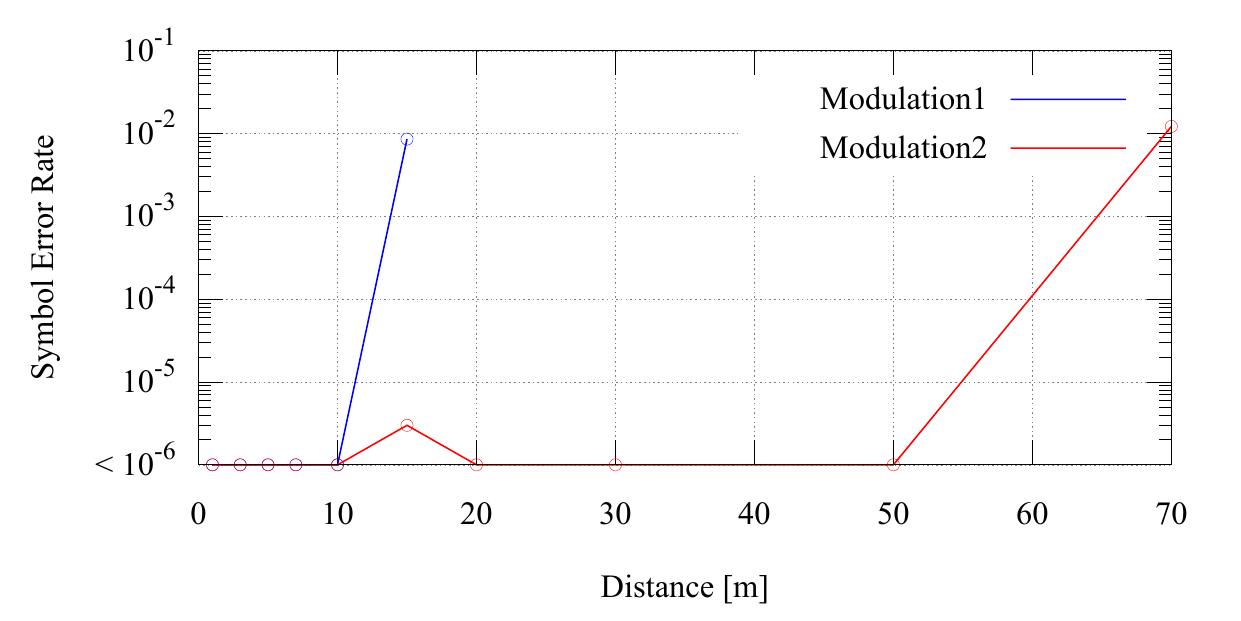}
	\caption{Symbol error rate for communication at each distance.}
	\label{fig:ber_results}
\end{figure}

We conducted indoor transmission experiments with the setup shown in Fig.~\ref{fig:exp}. Initially, we executed Modulation 1 and Modulation 2 at their optimized distances, respectively. The EVS output and demodulation results are presented in Fig.~\ref{fig:peak_exp}. The times between local maximums closely align with the symbols delineated in Table I, suggesting the feasibility of demodulation.
The local maximum of the acquired event is used to demodulate the event, as indicated by (\ref{eq:local_max})-(\ref{eq:select}).
The two EIM modulation methods set up were employed for communication over distances ranging from 0.2 m to 70 m.
Fig.~\ref{fig:ber_results} shows the BER by employing two types of EIM.

%

Since each symbol set in EIM has not been assigned a specific bit pattern in this paper, the performance is evaluated based on the symbol error rate (SER) rather than the bit error rate (BER). Although we cannot define a conventional BER-based FEC threshold commonly used in optical communication, we instead interpret the feasibility of transmission based on an SER below $10^{-4}$.
When a SER of $10^{-4}$, was permitted, propagation distance in an indoor environment employing Modulation 1 attained a range of 10 m at a data rate of 28 kbps, while Modulation 2 achieved a range of 50 m at 8.3 kbps under the same indoor conditions.
Successful transmission was achieved at longer distances than the 1 m and 10 m initially used for parameter tuning of each Modulation. This outcome reflects the robustness of the system design, which incorporated safety margins into the $T_r$ and $T_d$ settings as described in Sections~\ref{sec:tr} and \ref{sec:td}.

%

\section{Conclusion}\label{sec:cncl}
This paper proposed EIM, a novel modulation scheme designed specifically for event-based OCC systems. EIM exploits the time intervals between events to encode information, taking full advantage of the asynchronous and high-temporal-resolution nature of EVS. A key feature of EIM is its use of the negative events to explicitly identify pulse endings, allowing for precise interval calculation by referencing the timestamps of subsequent positive events. This approach enables efficient demodulation with minimal computational overhead.
By combining EIM with lightweight smoothing filters and simple interval extraction techniques, the proposed system achieves high-speed communication while preserving the low processing load that is a hallmark of EVS. Experimental results demonstrated transmission rates of 28 kbps over 10 m and 8.4 kbps over 50 m in an indoor environment, surpassing the bit rates of previously reported event-based OCC systems and setting a new benchmark.
These findings highlight the technical potential of EIM for real-time and resource-efficient OCC applications, particularly in environments such as indoor IoT networks or embedded vision systems. Future work focuses on extending the system for outdoor use, where robustness against burst errors and ambient light variation becomes essential. Designing error-resilient mechanisms compatible with the lightweight nature of EVS remains a key challenge moving forward.

\bibliographystyle{IEEEtran.bst}
\bibliography{bibliography}

\end{document}